\documentclass[conference]{IEEEtran}
\IEEEoverridecommandlockouts

\usepackage{cite}
\usepackage{amsmath,amssymb,amsfonts}
\usepackage{graphicx}
\usepackage{textcomp}
\usepackage{xcolor}
\usepackage{booktabs}
\usepackage{multirow}
\usepackage{array}
\usepackage{url}
\usepackage[hidelinks]{hyperref}
\usepackage{tabularx}

\newcommand{\method}{SambaGraph}
\newcommand{\safeimage}[2][]{%
\IfFileExists{image/#2}{\includegraphics[#1]{image/#2}}{%
\fbox{\parbox[c][1.25in][c]{0.92\linewidth}{\centering Missing figure file in the \texttt{image/} folder.}}}}

\def\BibTeX{{\rm B\kern-.05em{\sc i\kern-.025em b}\kern-.08em
    T\kern-.1667em\lower.7ex\hbox{E}\kern-.125emX}}

\begin{document}

\title{SambaGraph: Action--Reaction Spatio-Temporal Graphs for Soccer Tactical Response Modeling}

\author{
\IEEEauthorblockN{
Abel A. Reyes-Angulo\textsuperscript{1,4},
Henry O. Velesaca\textsuperscript{2,3,4},
and Steven Araujo\textsuperscript{3,4}
}
\IEEEauthorblockA{
\textsuperscript{1}Michigan Technological University, Houghton, MI, USA \quad
\textsuperscript{2}ESPOL Polytechnic University, Guayaquil, Ecuador
}
\IEEEauthorblockA{
\textsuperscript{3} University of Granada, Granada, Spain \quad
\textsuperscript{4}SambaSports AI, Guayaquil, Ecuador
}
\IEEEauthorblockA{
areyesan@mtu.edu; hvelesac@espol.edu.ec; saraujo@espol.edu.ec
}
}

\maketitle

\begin{abstract}
Soccer tactics are interactive: an attacking action changes the opponent's defensive problem, and the observed response depends on the multi-agent match state. We introduce \method, an action--reaction spatio-temporal graph dataset and benchmark for soccer tactical response modeling. From tracking and event data for all 64 matches of the 2022 FIFA World Cup, we curate 4,070 action-centered episodes represented as temporally aligned 23-node player--ball graph sequences with attack/defense views, response labels, and 26,270 split-safe attack--defense pairs. We study three questions: whether observed responses can be classified from graph episodes, whether successful defenses can be retrieved for a query attack, and whether graph-derived summaries support grounded LLM reasoning. A compact signature MLP obtains $0.796\pm0.007$ macro-F1 for response classification, while a fused graph--signature dual encoder reaches $0.471\pm0.029$ Hit@5 and $0.655\pm0.051$ Hit@10 for full-bank defensive retrieval. Hard negatives maximize pair discrimination but not retrieval quality. Local LLMs underperform supervised encoders for direct classification and do not improve over a strong original order in eight-candidate reranking, but they provide grounded tactical rationales. These results position \method\ as a reproducible benchmark for graph-based soccer strategy-response research. Code and dataset are available at: \url{https://github.com/areyesan/SambaGraph}.
\end{abstract}


\section{Introduction}
\label{sec:intro}

Soccer is a multi-agent visual reasoning problem. A pass, cross, shot, foul, or set piece immediately changes the opponent's defensive problem: defenders press, cover, clear, recover shape, or concede a chance. Although tracking and event data make these interactions observable, most public benchmarks focus on event recognition, localization, or outcome prediction rather than \emph{action--reaction} supervision: given an attacking context, what defensive response followed, and which past defensive examples are tactically relevant?

\begin{figure*}[t]
    \centering
    \safeimage[width=0.98\textwidth]{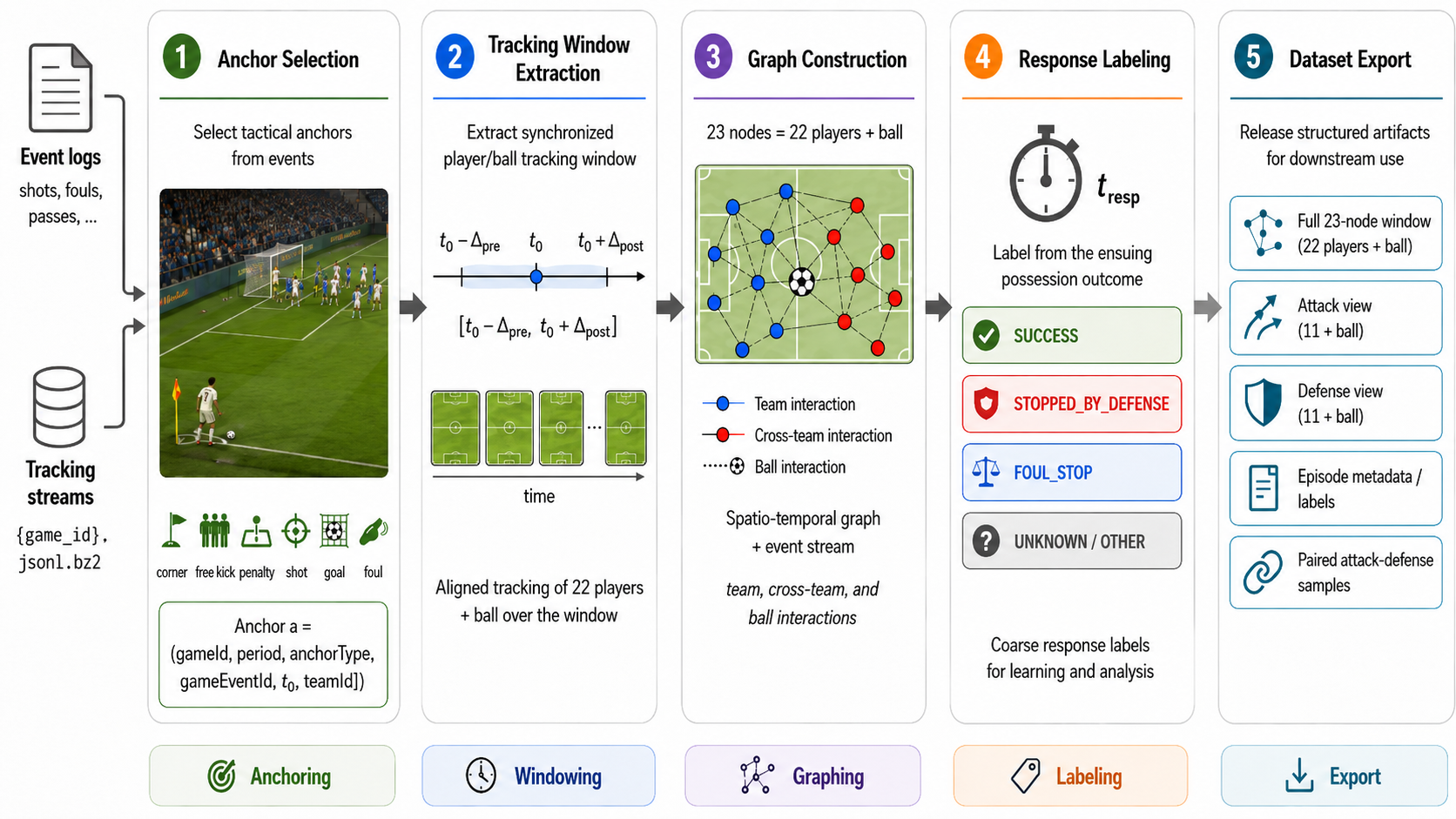}
    \caption{\method{} curation pipeline. Event and tracking streams are aligned through tactical anchors, converted into synchronized player--ball graph windows, labeled by the ensuing possession outcome, and exported as full graphs, team views, metadata, and paired samples.}
    \label{fig:sambagraph_pipeline}
\end{figure*}

This work introduces \method, a curated dataset and benchmark for tactical response modeling from tracking and event data. We use the PFF FC Enhanced 2022 World Cup data release, which provides match-level tracking, event logs, and metadata for all 64 matches \cite{pff_wc_release,kloppy_pff}. As Fig.~\ref{fig:sambagraph_pipeline} illustrates, we convert raw data into action-centered graph episodes: each episode starts from an on-ball anchor, aligns nearby tracking frames, constructs a fixed player--ball graph sequence, assigns a response label, and exports attack/defense views and paired examples for classification and retrieval.

Our contributions are threefold. First, we curate a graph-structured action--reaction dataset with 4,070 episodes from 64 World Cup matches. Second, we define response classification and attack-to-defense retrieval protocols with match-level split safety. Third, we report repeated-seed baselines showing that compact signatures are strong for classification, graph--signature fusion improves retrieval, hard-negative pair discrimination does not necessarily improve retrieval, and local LLMs are best used as explanation/reranking modules rather than primary predictors.
These contributions are organized around three research questions. 
\textbf{RQ1:} Can observed tactical responses be classified from action-centered player--ball graph episodes? 
\textbf{RQ2:} Can successful defensive responses be retrieved for a query attacking context? 
\textbf{RQ3:} Can graph-derived summaries support grounded LLM reasoning without replacing supervised graph/signature models?

\section{Related Work}
\label{sec:related}

Tracking data has enabled detailed modeling of team behavior, pass value, spatial control, and tactical decision support in soccer \cite{power2017passes,fernandez2021soccermap,le2017coordinated}. Large-scale resources such as SoccerNet have further accelerated soccer video understanding \cite{cioppa2022soccernetv3,soccernet_repo}. These works are valuable, but many benchmarks are centered on recognizing events or estimating value from a possession state. \method\ instead treats the tactical unit as an action--reaction episode: an attacking anchor, the surrounding multi-agent configuration, the observed defensive response, and retrieved examples that may help explain the response.

Graph representations are a natural fit for interacting players and the ball \cite{battaglia2018relational,kipf2017gcn,velickovic2018gat}, and spatio-temporal graph models have been successful for structured visual dynamics such as skeleton action recognition \cite{yan2018stgcn}. Recent sports-specific systems, including TacticAI for corner-kick analysis \cite{wang2024tacticai} and temporal graph models for pass receiver/outcome prediction \cite{rahimian2024pass}, show the value of graph-based tactical reasoning. Dynamic graph methods such as DyRep, JODIE, TGAT, EvolveGCN, and TGN model interactions that evolve over time \cite{trivedi2019dyrep,kumar2019jodie,xu2020tgat,pareja2020evolve,rossi2020tgn}. \method\ complements this literature by releasing broad action--response supervision across multiple event types and by evaluating both response prediction and example-based retrieval.

\section{Dataset Construction}
\label{sec:dataset}

\subsection{Source data and event anchors}
We derive \method\ from the PFF FC Enhanced 2022 World Cup dataset \cite{pff_wc_release}. We flatten event logs into a unified table containing timestamps, teams, players, set-piece types, possession events, shot outcomes, and foul outcomes. We select high-signal anchors: shots, goals, corners, free kicks, penalties, and explicit fouls. Each anchor is represented as
\begin{equation}
\begin{aligned}
a=(\texttt{gameId},\texttt{period},\texttt{anchorType},\texttt{gameEventId},\\
t_0,\texttt{teamId}).
\end{aligned}
\end{equation}
and anchors are deduplicated by event identifier and type.

\subsection{Tracking windows and graph tensors}
For an anchor at time $t_0$, we extract
\begin{equation}
W(a)=\{f_t\mid t\in[t_0-5\mathrm{s},t_0+10\mathrm{s}]\},
\end{equation}
canonicalize attack direction so the attacking team moves toward $+x$, and uniformly sample each episode to $L=64$ temporal steps for sequence baselines. This converts each 15-second window into a fixed 64-step tensor, corresponding to an effective sampled rate of approximately 4.3 Hz in the released representation. The task therefore models the observed action--reaction episode, not strict pre-anchor forecasting: the input may include early post-anchor defensive motion, while the label summarizes the subsequent possession outcome. 
Pre-anchor-only anticipation is a future benchmark variant.

At each sampled step $\ell$, the full view is a graph $G_\ell=(V_\ell,E_\ell,X_\ell)$ with 23 fixed nodes: 11 home players, 11 away players, and the ball. Node features include location, speed when available, team identity, and jersey identifiers. Let
\begin{equation}
X_\ell\in\mathbb{R}^{23\times d},\quad
A_\ell^{(r)}\in\{0,1\}^{23\times 23},\quad r\in\{\mathrm{team},\mathrm{opp},\mathrm{ball}\}
\end{equation}
be the node-feature matrix and relation-specific adjacency matrices for intra-team $k$NN, cross-team nearest-opponent, and ball-player edges. We use $k_{\mathrm{team}}=3$, $k_{\mathrm{opp}}=2$, and $k_{\mathrm{ball}}=5$, and define $A_\ell=\bigvee_r A_\ell^{(r)}$. Edge attributes include relative displacement, Euclidean distance, and relation type. The episode graph sequence is
\begin{equation}
\mathcal{G}(a)=\{(X_\ell,A_\ell,E_\ell)\}_{\ell=1}^{L},
\end{equation}
with synchronized 12-node attack and defense views formed by selecting one team plus the ball.

\subsection{Response labels and pair mining}
Each episode receives a fine response label from the event stream, including \texttt{GOAL}, \texttt{SHOT\_*}, \texttt{FOUL\_*}, and \texttt{END\_*}. We collapse long-tailed fine labels into four coarse tactical outcomes: \texttt{SUCCESS} for goals and high-value attacking outcomes, \texttt{STOPPED\_BY\_DEFENSE} for unsuccessful shots, clearances, interceptions, and end-of-attack events, \texttt{FOUL\_STOP} for foul-stopped possessions, and \texttt{UNKNOWN} for ambiguous or unresolved cases. This mapping preserves the tactical distinction needed for the benchmark while avoiding unstable fine-label classes. For retrieval, we construct split-safe pairs
\begin{equation}
(a_i,d_j,w_{ij},\texttt{pairType})\in\mathcal{P},
\end{equation}
where $a_i$ is a query attack and $d_j$ is a candidate defensive response. Pair types include same-episode positives, retrieved successful defensive positives, and hard failure negatives mined using a lightweight signature over ball trajectory and team-shape summaries. Tables~\ref{tab:validation}, \ref{tab:coarse_counts}, and \ref{tab:pairs} summarize scale, labels, and pair composition.

\begin{table}[t]
\caption{Dataset scale and validation summary.}
\label{tab:validation}
\centering
\begin{tabular}{lr}
\toprule
Item & Value \\
\midrule
Matches & 64 \\
Episodes & 4,070 \\
Attack--defense pairs & 26,270 \\
Automated validation checks & 20 \\
Failed validation checks & 0 \\
Full NPZ tensor coverage & 100\% \\
Attack-view NPZ coverage & 100\% \\
Defense-view NPZ coverage & 100\% \\
Media coverage & 100\% \\
Cross-split pair leakage & 0 \\
\bottomrule
\end{tabular}
\end{table}

\begin{table}[t]
\caption{Coarse response label distribution.}
\label{tab:coarse_counts}
\centering
\begin{tabular}{lr}
\toprule
Coarse label & Episodes \\
\midrule
\texttt{STOPPED\_BY\_DEFENSE} & 2,201 \\
\texttt{FOUL\_STOP} & 1,066 \\
\texttt{UNKNOWN} & 464 \\
\texttt{SUCCESS} & 339 \\
\bottomrule
\end{tabular}
\end{table}

\begin{table}[t]
\caption{Pair distribution for response matching.}
\label{tab:pairs}
\centering
\begin{tabular}{lr}
\toprule
Pair type & Pairs \\
\midrule
\texttt{pos\_same} & 4,070 \\
\texttt{pos\_retrieved\_success} & 9,990 \\
\texttt{neg\_hard\_failure} & 12,210 \\
\bottomrule
\end{tabular}
\end{table}

\subsection{Artifact schema}
The curated release contains global metadata and per-match artifacts keyed by \texttt{attack\_id}: flattened events, anchors, episode metadata, full 23-node NPZ windows, 12-node team views, response metadata, training pairs, compact episode contexts, and optional GIF/MP4 visualizations. This schema supports tabular, sequence, and graph baselines without reparsing the raw source files.

\section{Benchmark Tasks and Baselines}
\label{sec:tasks}

\subsection{Response classification}
Given an episode graph sequence, the goal is to classify the coarse observed response label. We report accuracy, balanced accuracy, macro-F1, and weighted-F1 over three seeds. Baselines are: (i) a compact MLP over episode-level tactical signatures, (ii) a dynamic $k$NN graph-GRU that applies frame-level graph message passing before temporal aggregation, and (iii) a graph--signature fusion model. This setup is response modeling over the full action--reaction window, not a claim of pre-anchor anticipation. This clarification is important because the post-anchor frames may contain part of the reaction being modeled; the benchmark is therefore designed to represent, classify, and retrieve observed tactical responses rather than forecast them from only pre-event context.

\subsection{Attack-to-defense retrieval and pair scoring}
Given a query attack, the model ranks candidate defensive responses. A relevant retrieval is a successful defensive response under the success-only protocol. We compare random retrieval, signature cosine retrieval, random/hard/mixed-negative dual encoders, and a fused graph--signature dual encoder. We report Hit@1, Hit@5, Hit@10, MRR, and nDCG@10. Full-bank retrieval ranks all eligible same-split candidate responses, which makes the task harder and more realistic than reranking a short list. We also evaluate pair classification with AUC and AP to test whether curated positives and negatives are separable; this diagnostic is intentionally distinct from full-bank retrieval because a model can separate pair labels without producing the best global ranking of successful defenses.

\subsection{Auxiliary LLM graph-summary reasoning}
We evaluate a lightweight language-based protocol in which each graph sequence is converted into a structured text summary with anchor type, ball displacement, team spread, compactness, and local pressure. LLMs either predict the coarse label or rerank eight candidate defenses. Candidate outcome labels are masked from the prompt and used only for evaluation, as shown in Fig.~\ref{fig:llm_prompt_example}. This protocol tests whether graph summaries support tactical reasoning; it is not treated as a replacement for tensor-based graph or signature models.

\begin{figure}[t]
\centering
\fbox{%
\begin{minipage}{0.96\columnwidth}
\small
\textbf{System.} You are a soccer tactics analyst and graph reasoning assistant. Compare a query attacking context with candidate defensive responses and return valid JSON.\\[2pt]
\textbf{User.} Rank the candidate defensive responses for the query attack. Prefer candidates that are tactically plausible and likely to correspond to a successful defensive response.\\
\texttt{Query attack: Anchor type = corner; ball displacement = (2.8, 6.6); attack compactness = 12.4; nearest player to ball = 3.1m; players within 10m = 5; ...}\\
\texttt{Candidate C1: Anchor type = free kick; defense compactness = 7.8; depth = 29.4; nearest player to ball = 1.2m; ...}\\
\texttt{Candidate C2: Anchor type = corner; defense compactness = 12.8; depth = 72.2; nearest player to ball = 6.5m; ...}\\
\textbf{Output JSON schema:} \texttt{\{"ranking":["C1","C2",...], "best":"C1", "rationale":"brief reason"\}}
\end{minipage}}
\caption{Example of the label-masked LLM reranking prompt built from graph-derived summaries.}
\label{fig:llm_prompt_example}
\end{figure}

\section{Experiments and Results}
\label{sec:results}

\subsection{Dataset validation}
Table~\ref{tab:validation} shows that all validation checks pass. In particular, all full/attack/defense NPZ files exist, sampled tensors load with expected node counts, pair identifiers join to existing episodes, media artifacts are present, and all curated pairs remain split-safe. These checks are important because leakage can otherwise occur when a retrieved defense comes from a match assigned to a different split than the query attack.

All experiments use match-level train/validation/test splits to prevent leakage across games. 
We report mean and standard deviation over three seeds for learned baselines. 
Classification is evaluated with accuracy, balanced accuracy, macro-F1, and weighted-F1, while retrieval is evaluated with Hit@K, MRR, and nDCG@10. 
The LLM experiments use the same test episodes but receive only label-masked graph-summary text; outcome labels are used only for evaluation.

\subsection{Response classification}
\begin{figure}[t]
\centering
\safeimage[width=0.98\linewidth]{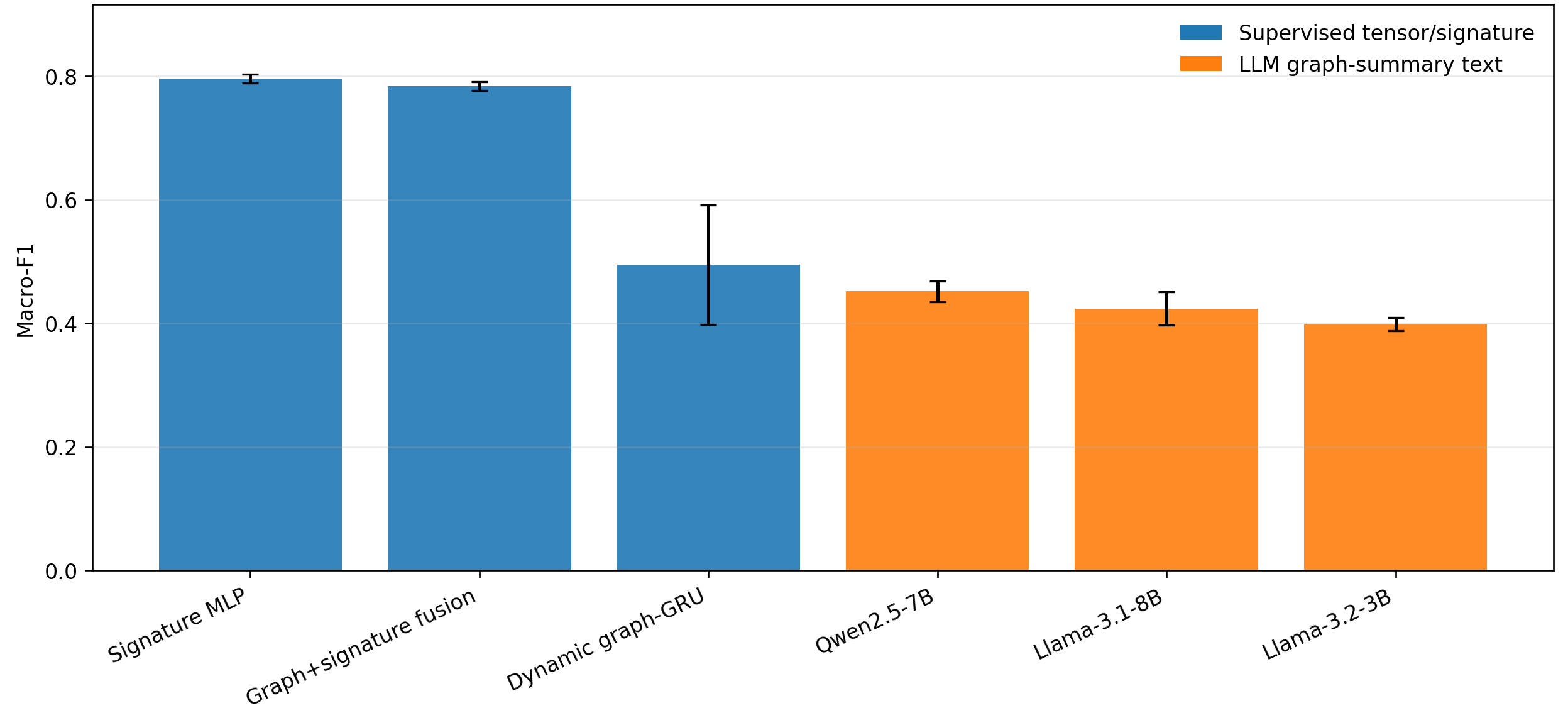}
\caption{Coarse response classification macro-F1. Supervised models outperform LLMs that receive only graph-summary text.}
\label{fig:combined_classification_macro_f1}
\end{figure}

Table~\ref{tab:classification_combined} reports coarse response classification. The signature MLP is strongest, reaching $0.842\pm0.009$ accuracy and $0.796\pm0.007$ macro-F1. Graph--signature fusion is close, while the graph-only GRU is lower and higher-variance. Local LLM classifiers are weaker than supervised encoders: Qwen2.5-7B obtains the best LLM macro-F1 ($0.451\pm0.017$). Fig.~\ref{fig:combined_classification_macro_f1} visualizes the same trend and supports the interpretation that graph-derived summaries contain useful signal but do not replace specialized supervised encoders. The fact that the signature MLP remains strongest also provides a useful benchmark constraint: future graph models should be compared against compact tactical features rather than only against weak baselines.

\begin{table*}[t]
\caption{Coarse response classification. Supervised models use tensors/signatures; LLMs use graph-summary text. Results are mean $\pm$ standard deviation over three seeds.}
\label{tab:classification_combined}
\centering
\resizebox{\textwidth}{!}{%
\begin{tabular}{lllrrr}
\toprule
Model & Input & Protocol & Acc. & Bal. Acc. & Macro-F1 \\
\midrule
Signature MLP & Signature vector & Supervised & $0.842\pm0.009$ & $0.825\pm0.004$ & $0.796\pm0.007$ \\
Graph--signature fusion & Graph + signature & Supervised & $0.842\pm0.001$ & $0.788\pm0.013$ & $0.784\pm0.007$ \\
Dynamic graph-GRU & Graph tensor & Supervised & $0.543\pm0.087$ & $0.538\pm0.098$ & $0.495\pm0.097$ \\
\midrule
Qwen2.5-7B & Graph-summary text & LLM zero-shot & $0.504\pm0.022$ & $0.448\pm0.017$ & $0.451\pm0.017$ \\
Llama-3.1-8B & Graph-summary text & LLM zero-shot & $0.527\pm0.040$ & $0.469\pm0.031$ & $0.424\pm0.027$ \\
Llama-3.2-3B & Graph-summary text & LLM zero-shot & $0.460\pm0.012$ & $0.482\pm0.012$ & $0.398\pm0.011$ \\
\bottomrule
\end{tabular}}
\end{table*}

\subsection{Attack-to-defense retrieval and reranking}
\begin{figure}[t]
\centering
\safeimage[width=0.98\columnwidth]{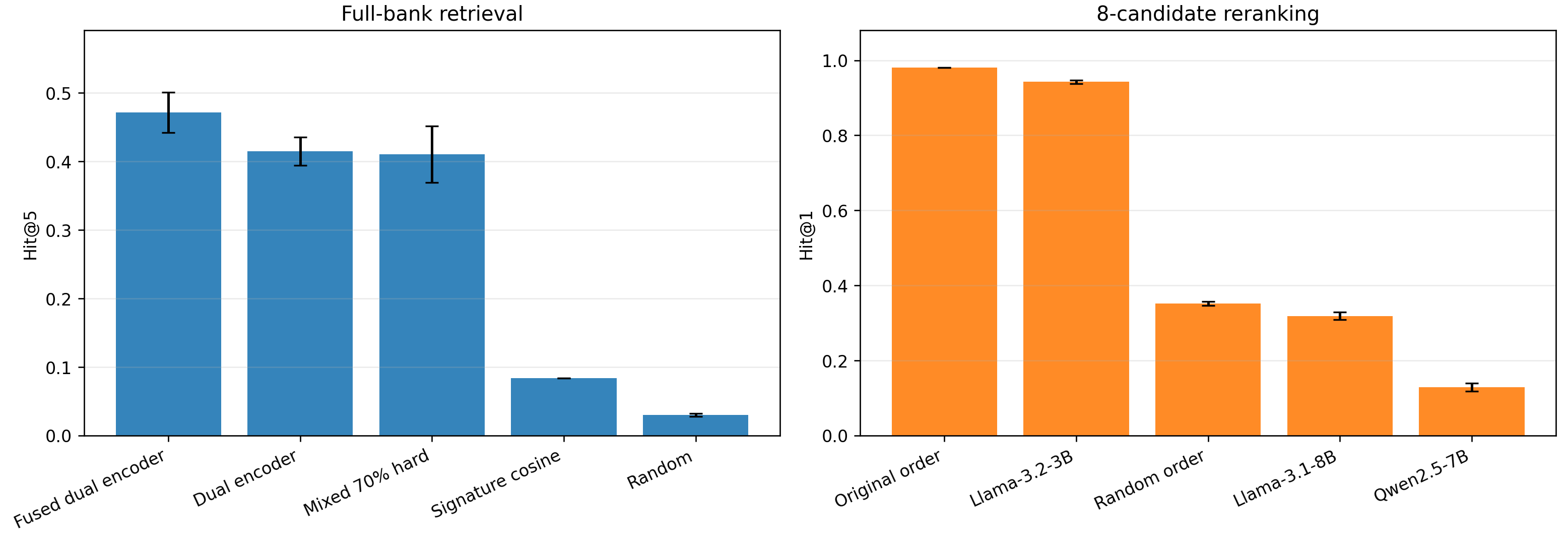}
\caption{Full-bank retrieval and eight-candidate reranking. The fused graph--signature encoder is strongest for full-bank retrieval; Llama-3.2-3B best preserves the strong original reranking order.}
\label{fig:combined_retrieval_reranking}
\end{figure}

Table~\ref{tab:retrieval_reranking_combined} compares full-bank retrieval with the separate eight-candidate LLM reranking protocol. These protocols are not equivalent: full-bank models rank the same-split retrieval pool, whereas LLMs rerank eight preselected candidates with outcome labels hidden. In full-bank retrieval, the fused graph--signature dual encoder is best, reaching $0.471\pm0.029$ Hit@5 and $0.655\pm0.051$ Hit@10. In reranking, the original candidate order is already very strong ($0.980\pm0.000$ Hit@1); Llama-3.2-3B largely preserves this order ($0.943\pm0.005$ Hit@1), while larger local models degrade top-ranked success. Fig.~\ref{fig:combined_retrieval_reranking} shows the protocol-specific results side by side.

\begin{table*}[t]
\caption{Retrieval and reranking results. Full-bank retrieval searches same-split candidates; LLMs rerank eight preselected defenses with labels hidden.}
\label{tab:retrieval_reranking_combined}
\centering
\resizebox{\textwidth}{!}{%
\begin{tabular}{llrrrrr}
\toprule
Model & Protocol & Hit@1 & Hit@5 & Hit@10 & MRR & nDCG@10 \\
\midrule
Fused dual encoder & Full-bank retrieval & $0.193\pm0.014$ & $0.471\pm0.029$ & $0.655\pm0.051$ & $0.328\pm0.020$ & $0.295\pm0.007$ \\
Dual encoder & Full-bank retrieval & $0.164\pm0.018$ & $0.414\pm0.021$ & $0.579\pm0.041$ & $0.293\pm0.021$ & $0.247\pm0.023$ \\
Mixed 70\% hard & Full-bank retrieval & $0.138\pm0.031$ & $0.411\pm0.041$ & $0.588\pm0.004$ & $0.275\pm0.029$ & $0.235\pm0.024$ \\
Signature cosine & Full-bank retrieval & $0.018\pm0.000$ & $0.083\pm0.000$ & $0.122\pm0.000$ & $0.052\pm0.000$ & $0.039\pm0.000$ \\
Random & Full-bank retrieval & $0.006\pm0.002$ & $0.030\pm0.002$ & $0.061\pm0.003$ & $0.033\pm0.002$ & $0.014\pm0.001$ \\
\midrule
Original order & 8-candidate reranking & $0.980\pm0.000$ & $0.981\pm0.002$ & $1.000\pm0.000$ & $0.983\pm0.000$ & $0.986\pm0.000$ \\
Llama-3.2-3B & 8-candidate reranking & $0.943\pm0.005$ & $0.977\pm0.002$ & $1.000\pm0.000$ & $0.954\pm0.003$ & $0.959\pm0.003$ \\
Random order & 8-candidate reranking & $0.352\pm0.005$ & $0.959\pm0.001$ & $1.000\pm0.000$ & $0.584\pm0.004$ & $0.677\pm0.002$ \\
Llama-3.1-8B & 8-candidate reranking & $0.319\pm0.010$ & $0.821\pm0.010$ & $1.000\pm0.000$ & $0.539\pm0.007$ & $0.694\pm0.003$ \\
Qwen2.5-7B & 8-candidate reranking & $0.128\pm0.011$ & $0.801\pm0.008$ & $1.000\pm0.000$ & $0.388\pm0.008$ & $0.590\pm0.005$ \\
\bottomrule
\end{tabular}}
\end{table*}

\subsection{Negative mining and LLM interpretation}
\begin{figure}[t]
\centering
\safeimage[width=1.0\columnwidth]{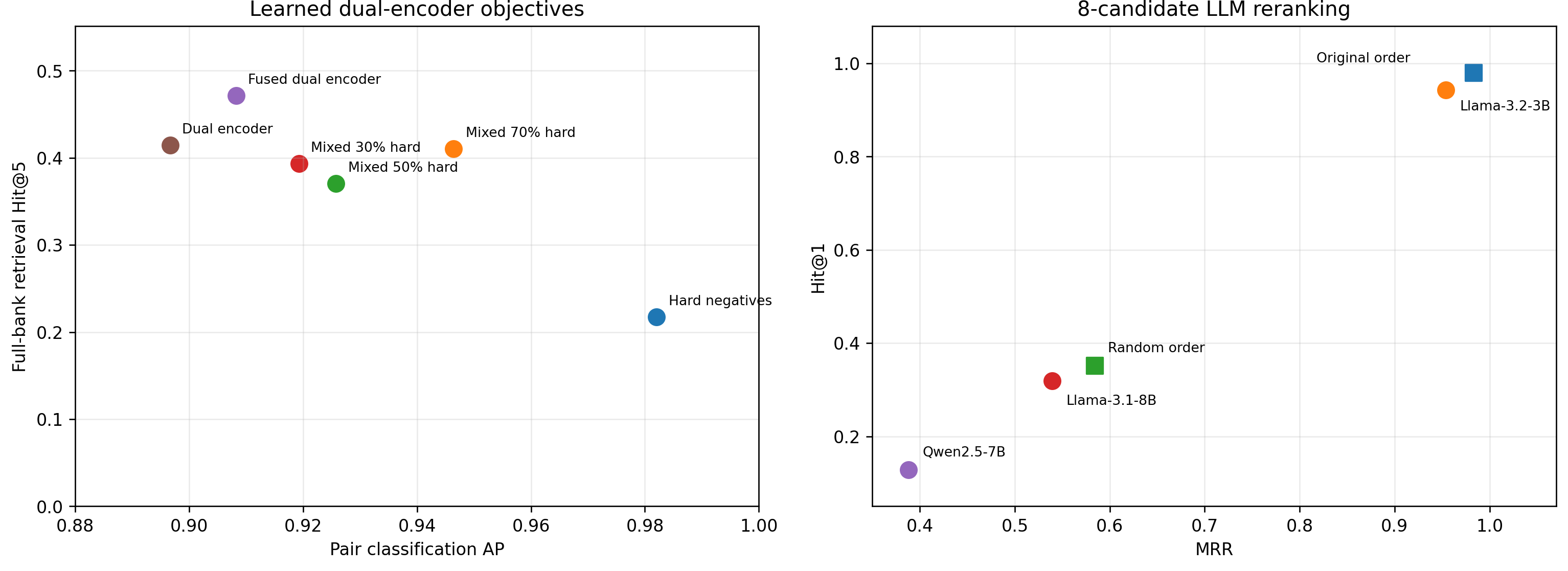}
\caption{Comparison of pair discrimination, full-bank retrieval, and eight-candidate LLM reranking. Hard negatives improve pair AP, but retrieval and reranking quality follow different trends.}
\label{fig:pair_discrimination_retrieval_llm_combined}
\end{figure}

Table~\ref{tab:pair_classification_mean_std} reports the pair-discrimination sanity check. Hard-negative training gives the best pair AUC/AP, but Fig.~\ref{fig:pair_discrimination_retrieval_llm_combined} shows that this does not translate monotonically to full-bank retrieval. The result is methodological: pair classification tests separability of curated positives and negatives, whereas retrieval tests whether successful defenses are ranked well among many candidates.

\begin{table}[t]
\caption{Pair classification on curated attack--defense pairs.}
\label{tab:pair_classification_mean_std}
\centering
\resizebox{\columnwidth}{!}{%
\begin{tabular}{lrr}
\toprule
Model & AUC & AP \\
\midrule
Dual encoder, hard negatives & $0.983\pm0.014$ & $0.982\pm0.014$ \\
Dual encoder, mixed 70\% hard & $0.947\pm0.008$ & $0.946\pm0.002$ \\
Dual encoder, mixed 50\% hard & $0.928\pm0.005$ & $0.926\pm0.006$ \\
Dual encoder, mixed 30\% hard & $0.924\pm0.006$ & $0.919\pm0.003$ \\
Fused dual encoder, random negatives & $0.900\pm0.015$ & $0.908\pm0.013$ \\
Dual encoder, random negatives & $0.903\pm0.006$ & $0.897\pm0.007$ \\
\bottomrule
\end{tabular}}
\end{table}

\begin{figure}[h]
\centering
\safeimage[width=0.96\linewidth]{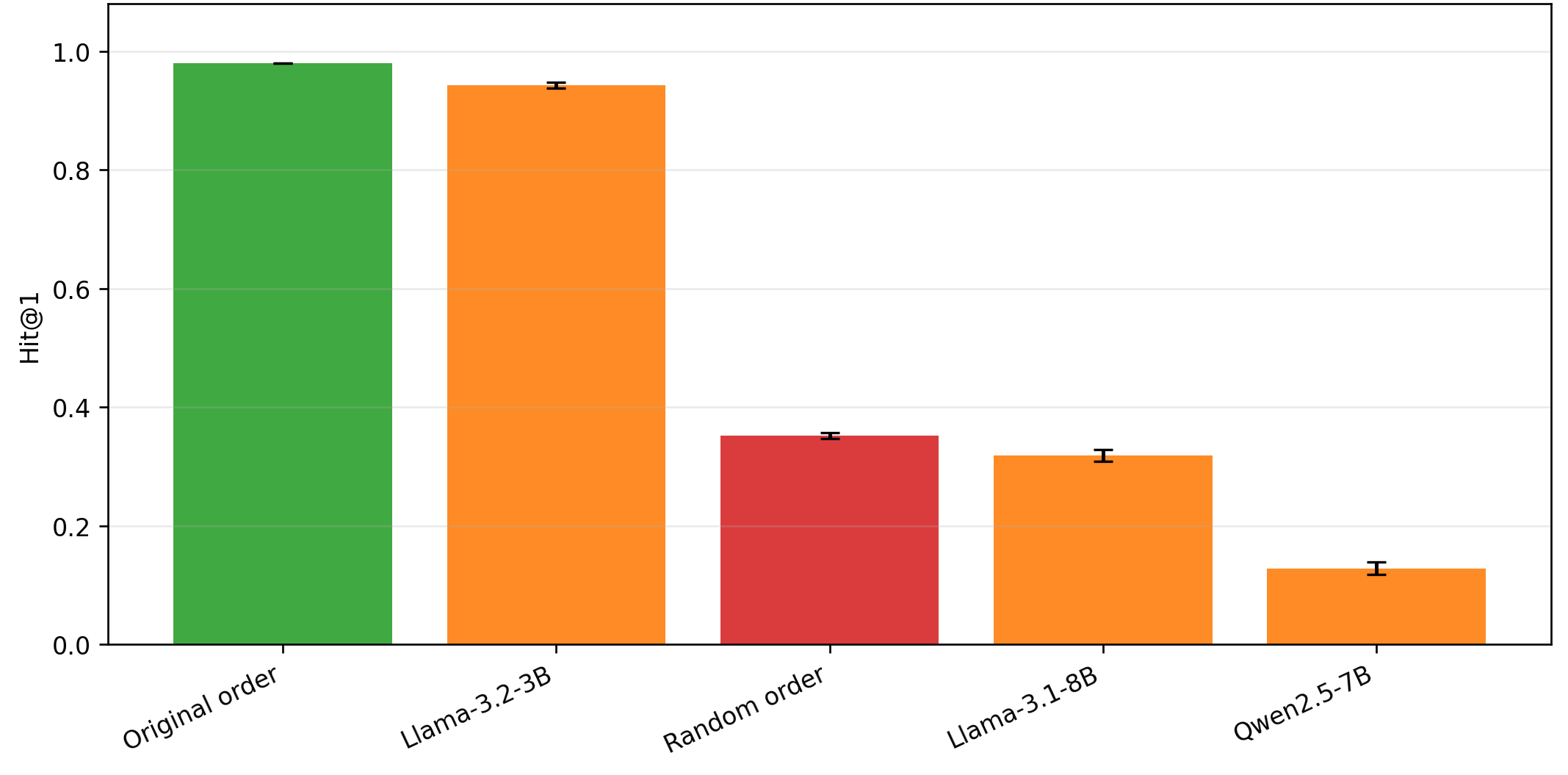}
\caption{Eight-candidate LLM reranking with outcome labels masked. The original order is strongest; Llama-3.2-3B preserves it better than the other local LLMs.}
\label{fig:llm_rerank_with_baselines_clean}
\end{figure}

The LLM results add a complementary perspective. The LLM classifier is not competitive with supervised numerical encoders, and LLM reranking does not beat the original candidate order. However, Llama-3.2-3B preserves most of that ordering and produces natural-language tactical rationales from label-masked graph summaries. Thus, we treat LLMs as explanation/reranking modules rather than primary graph encoders. Fig.~\ref{fig:llm_rerank_with_baselines_clean} reports the corresponding Hit@1 comparison.

\begin{figure*}[h]
\centering
\safeimage[width=0.98\linewidth]{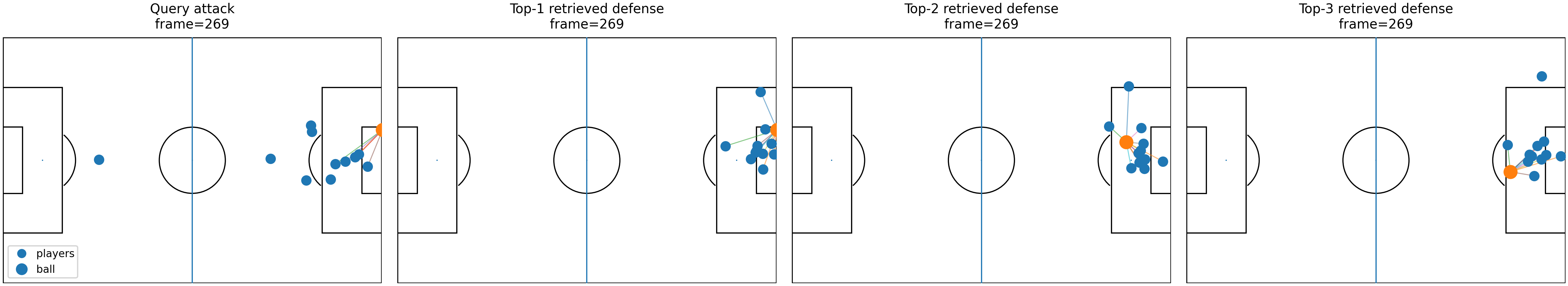}
\caption{Qualitative retrieval example showing a query attack and retrieved defensive response candidates rendered on the pitch.}
\label{fig:qualitative}
\end{figure*}

\subsection{Qualitative retrieval}
Beyond aggregate metrics, retrieval outputs are useful only if they are spatially interpretable.
Fig.~\ref{fig:qualitative} illustrates how retrieved examples can be inspected on the pitch. These qualitative views are not used to compute the metrics, but they are useful for auditing whether retrieved defenses are not only label-correct but also spatially and tactically plausible. This is important for sports analytics applications, where a recommendation should be interpretable to analysts and coaches.

\section{Discussion}
\label{sec:discussion}

The experiments answer the motivating questions. First, tracking and event logs can be transformed into validated action--reaction graph episodes with split-safe labels and retrieval supervision. Second, observed response classification is feasible, but compact signatures remain difficult to beat; this weakens any claim that graph encoders alone drive classification performance and establishes a strong baseline for future graph models. This outcome is expected in a limited-data setting: hand-crafted signatures directly encode ball displacement, team spread, compactness, and local pressure, whereas graph encoders must learn these abstractions from relatively few episodes. Third, graph information is useful for the retrieval use case: the fused graph--signature dual encoder is the strongest full-bank retrieval model. This retrieval setting is closest to the intended analytics use case: given a current attacking configuration, an analyst can inspect similar historical defensive responses and compare how different teams contained or failed to contain comparable situations. Fourth, negative mining changes what the embedding learns, with hard negatives improving pair AP but reducing broad success-oriented retrieval. Finally, local LLMs can reason over graph-derived summaries, but their practical role is currently grounded explanation and short-list reranking, not direct prediction.

The benchmark should therefore be read primarily as a dataset and protocol contribution rather than as a new architecture paper. This is intentional. By providing the graph tensor representation, response-label mapping, pair construction, validation checks, and multiple baseline families, \method\ makes it possible to study when relational structure helps, when compact tactical descriptors are sufficient, and how retrieval can support interpretable soccer analysis.

\section{Limitations and Ethics}
\label{sec:limitations}

\method\ models observed action--reaction outcomes rather than optimal tactical decisions. 
It is also not a pure pre-anchor anticipation benchmark, because the released full-window representation includes early post-anchor motion.
A successful response may depend on factors not fully captured in the graph, such as attacker error, goalkeeper performance, score state, fatigue, or coaching instructions. 
Coarse labels compress rich tactical behavior, and broadcast tracking may include occlusions, identity switches, or positional noise. 
The benchmark is intended for aggregate tactical research, not individual player assessment. 
Any release must respect the source-data license; if raw tracking/event data cannot be redistributed, the public release should provide curation code, schemas, manifests, validation notebooks, and benchmark scripts while requiring users to obtain the original data through the official source. 

\section{Conclusion}
\label{sec:conclusion}

We introduced \method, a curated action--reaction graph dataset and benchmark for soccer tactical response modeling. The dataset contains 4,070 validated episodes from 64 World Cup matches, synchronized attack/defense graph views, and 26,270 curated pairs. The baselines show that compact signatures are strong for response classification, graph--signature fusion improves full-bank defensive retrieval, and pair discrimination is not equivalent to retrieval quality. Multi-seed local LLM experiments show that graph summaries support grounded rationales, but current LLMs remain below supervised encoders for classification and do not improve over a strong original reranking order. Together, these results establish \method\ as a practical benchmark for graph reasoning and strategy-response research in soccer.

%\newpage

\end{document}